\documentclass[11pt,a4paper]{article}
\usepackage[hyperref]{emnlp2020}
\RequirePackage{xcolor}
\usepackage{times}
\usepackage{tikz}
\usetikzlibrary{positioning}
\usetikzlibrary{arrows}
\usepackage{latexsym,amsmath,amssymb,bm}

\usepackage{graphicx}
\usepackage{microtype}
\RequirePackage{soul}
\RequirePackage[skip=.5ex]{caption}
\RequirePackage[skip=.4ex]{subcaption}
\RequirePackage[textsize=scriptsize]{todonotes}
\RequirePackage[inline]{enumitem}
\setlist[itemize]{nosep,leftmargin=*,labelwidth=0pt}
\setlist[enumerate]{nosep}
\setlist[description]{nosep,leftmargin=.8em}
\RequirePackage{wrapfig}
\makeatletter
\g@addto@macro{\normalsize}{%
\setlength{\abovedisplayskip}{0pt}%
\setlength{\abovedisplayshortskip}{0pt}%
\setlength{\belowdisplayskip}{0pt}%
\setlength{\belowdisplayshortskip}{0pt}}
\makeatother
\intextsep \textfloatsep

\aclfinalcopy 
\def\aclpaperid{2976} 

\newcommand\our{KYC}
\def\longname{Know Your Client}

\newcommand{\sunita}[1]{\textcolor{red}{#1}}

\begin{document}

\def\xtitle{NLP Service APIs and Models for Efficient Registration of New Clients}

\title{\xtitle}


\author{Sahil Shah$^1$, Vihari Piratla$^1$, Soumen Chakrabarti$^1$, Sunita Sarawagi$^1$ \\
$^1$Department of Computer Science, Indian Institute of Technology, Bombay, India\\
\texttt{sahilshah00199@gmail.com, \{vihari,soumen,sunita\}@cse.iitb.ac.in}}

\date{}
\maketitle

\begin{abstract}
State-of-the-art NLP inference uses enormous neural architectures and models trained for GPU-months, well beyond the reach of most consumers of NLP.  This has led to one-size-fits-all public API-based NLP service models by major AI companies, serving large numbers of clients.  Neither (hardware deficient) clients nor (heavily subscribed) servers can afford traditional fine tuning.  Many clients own little or no labeled data.  We initiate a study of adaptation of centralized NLP services to clients, and present one practical and lightweight approach.  Each client uses an unsupervised, corpus-based sketch to register to the service.  The server uses an auxiliary network to map the sketch to an abstract vector representation, which then informs the main labeling network.  When a new client registers with its sketch, it gets immediate accuracy benefits.  We demonstrate the success of the proposed architecture using sentiment labeling, NER, and predictive language modeling.
\end{abstract}


\newcommand{\lastE}{M}
\section{Introduction}
\label{sec:Intro}

State-of-the-art NLP uses large neural networks with billions of parameters, enormous training data, and intensive optimization over weeks of GPU-time, causing more carbon emission than a car over its lifetime~\citep{strubell2019energy}.  Such training prowess is (mercifully) out of reach for most users of NLP methods.  Recognizing this, large AI companies have launched NLP cloud services\footnote{\href{https://cloud.google.com/natural-language}{Google NLP}, \href{https://azure.microsoft.com/en-us/services/cognitive-services/language-understanding-intelligent-service/}{Microsoft Azure}, \href{https://www.ibm.com/watson/natural-language-processing}{IBM Watson}} and also provided trained models for download and fine tuning.  But many clients have too little data or hardware for fine tuning massive networks.  Neither can the service be expected to fine-tune for each client.

Distributional mismatch between the giant general-purpose corpus used to train the central service and the corpus from which a client's instances arise leads to lower accuracy.  A common source of trouble is mismatch of word salience \citep{paik2013novel} between client and server corpora~\cite{Ruder2019Neural}.  
%
In this respect, our setting also presents a new opportunity.  Clients are numerous and form natural clusters, e.g., healthcare, sports, politics.  We want the service to exploit commonalities in existing client clusters, without explicitly supervising this space, and provide some level of generalization to new clients without re-training or fine-tuning.  

In response to the above challenges and constraints, we initiate an investigation of practical protocols for lightweight client adaptation of NLP services.  We propose a system, \our~(``\longname''), in which each client registers with the service using a simple sketch derived from its (unlabeled) corpus.
The service network takes the sketch as additional input with each instance later submitted by the client.  The service provides accuracy benefits to new clients immediately.

What form can a client sketch take? How should the service network incorporate it?  While this will depend on the task, we initiate a study of these twin problems focused on predictive language modeling, sentiment labeling, and named entity recognition (NER).  We show that a simple late-stage  intervention in the server network gives visible accuracy benefits, and provide diagnostic analyses and insights.  Our code and data can be found {\href{https://github.com/sahil00199/KYC}{here}}\footnote{\url{https://github.com/sahil00199/KYC}}.

\paragraph{Contributions}
In summary, we
\begin{itemize}
\item introduce the on-the-fly client adaptation problem motivated by networked NLP API services;
\item present \our, that {\em learns to compute} client-specific biases from unlabeled client sketches; 
\item show improved accuracy for predictive language modeling, NER and sentiment labeling;
\item diagnose why \our's simple client-specific label biases succeed, in terms of relations between word salience, 
instance length and label distributions at diverse clients.
\end{itemize}

\paragraph{Related work}
Our method addresses the mismatch between a client's data distribution and the server model. The extensive domain adaptation literature~\cite{Daume2007,BlitzerMP06,Ben-David:2006} is driven by the same goal but most of these update model parameters using labeled or unlabeled data from the target domain (client). Unsupervised Domain Adaptation summarized in \cite{unsupervised_domain_adaptation} relaxes the requirement of labelled client data, but still demands target-specific fine-tuning which inhibits scalability. Some recent approaches attempt to make the adaptation light-weight \cite{LinL2018,LiSS2020,JiaLZ2019,CaiW2019,LiuWF2020} while  others propose to use entity description~\cite{BapnaTH2017,ShahGF2019} for zero-shot adaptation.  Domain generalization is another relevant technique~\cite{ChenC2018,GuoSB18,Li2018DomainGW,WangZZ2019,VihariSSS18,CarlucciAS2019,DouCK19,VihariNS2020} where multiple domains during training are used to train a model that can generalize to new domains.  Of these, the method that seems most relevant to our setting is the mixture of experts network of \cite{GuoSB18}, with which we present empirical comparison.
Another option is to transform the client data style so as to match the data distribution used to train the server model.  Existing style transfer techniques~\cite{YangZC2018, ShenLB2017,PrabhumoyeTS2018, FuTP2018,lample2018multipleattribute,LiJH2018,GongBW2019} require access to server data distribution.  

\section{Proposed service protocol}
\label{sec:Proposal}
We formalize the constraints on the server and client in the API setting.
\begin{enumerate*}[label=(\arabic*)]
\item The server is expected to scale to a large number of clients making it impractical to adapt to individual clients.
\item After registration, the server is expected to provide labeling immediately and response latency per instance must be kept low implying that the server's inference network cannot be too compute-intensive.
\item Finally,
the client cannot perform complex pre-processing of every instance before sending to the server, and does not have any labelled data.
\end{enumerate*}

\paragraph{Server network and model}
These constraints lead us to design a server model that \emph{learns to   compute} client-specific model parameters from the client sketch, and requires no client-specific fine-tuning or parameter learning.
The original server network is written as $\hat{\bm{y}} = Y_\theta(\lastE_\theta(\bm{x}))$ where $\bm{x}$ is the input instance, and
$Y_\theta$ is a softmax layer to get the predicted label $\hat{\bm{y}}$.
$\lastE_\theta$ is a representation learning layer that may take diverse forms depending on the task; of late, BERT \citep{devlin2018bert} is used to design $\lastE_\theta$ for many tasks.

We augment the server network to accept, with
\begin{wrapfigure}{r}{0.3\hsize}
\begin{tikzpicture}[>=stealth',align=center]
\node (jl) {loss};
\node [left=3mm of jl] (y) {$\bm{y}$};
\draw [->] (y) to (jl);
\node [rectangle, draw,
minimum width=9mm, below=3mm of jl] (Stheta) {$Y_\theta$};
\draw [->] (Stheta) to (jl);
\node [circle,draw, fill=yellow!15, 
inner sep=.1mm, below=2mm of Stheta] (plus) {$+$};
\draw [->] (plus) to (Stheta);
\node [rectangle, draw, anchor=center,
minimum width=9mm, minimum height=8mm,
below left=6mm and 2mm of plus] (Etheta) {$\lastE_\theta$};
\node [anchor=center, fill=yellow!15,
below right=3mm and 2mm of plus,
outer sep=.1mm, inner sep=0mm] (g) {$\bm{g}$};
\draw [->] (Etheta) to (plus);
\draw [->] (g) to (plus);
\node [rectangle, draw, anchor=center, fill=yellow!15,
minimum width=9mm, minimum height=8mm,
below=3mm of g] (Gphi) {$G_\phi$};
\draw [->] (Gphi) to (g);
\node [anchor=center,below=6mm of Etheta] (x) {$\bm{x}$};
\node [fill=yellow!15,
anchor=center,below=3mm of Gphi] (D) {$S_c$};
\draw [->] (x) to (Etheta);
\draw [->] (D) to (Gphi);
\draw [->, dotted] (D) to (x);
\end{tikzpicture}
\caption{\raggedright \our{} overview.}  \label{fig:overview}
\end{wrapfigure}
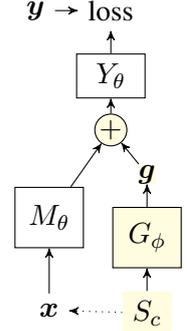
each input $\bm{x}$, a client-specific sketch~$S_c$ as shown in \figurename~\ref{fig:overview}.     
We discuss possible forms of $S_c$ in the next subsection.  (The dotted arrow represents a generative influence of $S_c$ on $\bm{x}$.)  The server implements an auxiliary network $\bm{g} = G_\phi(S_c)$.  Here $\bm{g}$ can be regarded as a neural digest of the client sketch.
Module $\bigoplus$ combines $\lastE_\theta(\bm{x})$ and $\bm{g}$; concatenation was found adequate on the tasks we evaluated but we also discuss other options in Section~\ref{sec:Expt}.  
When the $\bigoplus$ module is concatenation we are computing a client-specific per-label bias, and even that provides significant gains, as we show in Section~\ref{sec:Expt}.


\paragraph{Client sketch}
The design space of client sketch $S_c$ is infinite.  We initiate a study of designing $S_c$ from the perspective of term weighting and salience in Information Retrieval \citep{paik2013novel}.  $S_c$ needs to be computed once by each client, and thereafter reused with 
every input instance~$\bm{x}$.  
Ideally, $S_c$ and $G_\phi$ should be locality preserving, in the sense that clients with similar corpora and tasks should lead to similar~$\bm{g}$s. Suppose the set of clients already registered is~$C$.

A simple client sketch is just a vector of counts of all words in the client corpus.  Suppose word $w$ occurs $n_{c,w}$ times in a client $c$, with $\sum_w n_{c,w}=N_c$.  Before input to $G_\phi$, the server normalizes these counts using counts of other clients as follows:
From all of $C$, the server will estimate a background unigram rate of word. 
Let the estimated rate for word $w$ be~$p_w$, which is calculated as:
\begin{align}
p_w &= \textstyle (\sum_{c\in C}n_{c,w})\left/\left(\sum_w \sum_{c\in C}n_{c,w}\right)\right..
\end{align}
The input into $G_\phi$ will encode, for each word $w$, how far the occurrence rate of $w$ for client $c$ deviates from the global estimate.  Assuming the multinomial word event distribution, the marginal probability of having $w$ occur $n_{c, w}$ times at client $c$ is proportional to $p_w^{n_{c,w}} (1 - p_w)^{(N_c - n_{c,w})}$.  We finally pass a vector containing the normalized negative log probabilities as input to the model:
\begin{multline}
\label{eq:salience}
S_c \propto \Bigl( - n_{c,w}\log p_w \\[-2ex]
- (N_c - n_{c,w})\log(1-p_w): \forall w \Bigr).
\end{multline}
We call this the {\bf term-saliency} sketch.  We discuss other sketches like TF-IDF and corpus-level statistics like average instance length in Sec.~\ref{sec:ablation}.  

\section{Experiments}
\label{sec:Expt}
We evaluate \our\ on three NLP tasks as services: NER, sentiment classification, and auto-completion based on predictive language modeling. 
We compare \our\ against the baseline model (without the $G_\phi$ network in \figurename~\ref{fig:overview}) and the mixture of experts (MoE) model \citep{GuoSB18} (see Appendix B).  For all three models, the $\lastE_\theta$ network is identical in structure.  
In \our,  $G_\phi$ has two linear layers with ReLU giving a 128-dim vector~$\bm{g}$, with slight exceptions (see Appendix A). 
We choose datasets that are partitioned naturally across domains, used to simulate clients.  We evaluate in two settings: in-distribution (ID) on test instances from clients seen during training, and out-of-distribution (OOD) on instances from unseen clients.  For this, we perform a leave-k-client-out evaluation where given a set $D$ of clients, we remove $k$ clients as OOD test and use remaining $D - k$ as the training client set $C$. 

%
\noindent{\bf Named Entity Recognition (NER)} We use Ontonotes \cite{ontonotes} which has 18 entity classes 
from 31 sources which forms our set $D$ of clients.  We perform leave-2-out test five times with 29 training clients as $C$. 
We train a cased BERT-based NER model~\cite{devlin2018bert} 
and report F-scores.   
%
\setlength\tabcolsep{2.0pt}
\begin{table}
\centering
\begin{tabular}{|l | c c c | c c c |}
\hline
& \multicolumn{3}{|c|}{\small{OOD}} & \multicolumn{3}{|c|}{\small{ID}}  \\
OOD Clients & \small{Base} & \small{MoE} & \small{{\our}} & \small{Base}& \small{MoE} & \small{{\our}} \\
\hline
\small{BC/CCTV+Phoenix} & 63.8 & 66.9 & 71.8 & 86.0 & 83.8 & 86.7 \\
\small{BN/PRI+BN/VOA} & 88.7 & 87.9 & 90.7 & 84.5 & 83.0 & 86.0\\
\small{NW/WSJ+Xinhua} &73.9 & 78.9 & 80.9 & 80.8 & 77.2 & 82.5\\
\small{BC/CNN+TC/CH} & 78.3 & 75.2 & 78.7 & 85.6 & 82.7 & 87.4\\
\small{WB/Eng+WB/a2e} & 76.2 & 69.9 & 78.4 & 86.4 & 82.6 & 87.3 \\
\hline
Average & 76.2 & 75.8 & \textbf{80.1} & 84.7 & 81.9 & \textbf{86.0}\\
\hline
\end{tabular}
\vspace{0.3em}
\caption{Test F1 on Ontonotes NER. OOD numbers are on the two listed domains whereas ID numbers are on test data of clients seen during training.}
\label{table:ner_numbers}
\end{table}
Table \ref{table:ner_numbers} shows that \our\ provides substantial gains for OOD clients. For the first two OOD clients (BC/CCTV,Phoenix), the baseline F1 score jumps from 63.8 to 71.8. MoE performs worse than baseline. We conjecture this is because separate softmax parameters over the large NER label space is not efficiently learnable. 
    
\noindent{\bf Sentiment Classification} 
We use the popular Amazon dataset \cite{amazon_dataset} with each product genre simulating a client. 
We retain genres with more than 1000 positive and negative reviews each and randomly sample 1000 positive and negative reviews from these 22 genres. We perform leave-2-out evaluation five times and Table~\ref{table:sentiment_numbers} shows the five OOD genre pairs. 
We use an uncased BERT model for classifcation~\cite{sentiment_repo}. 
\begin{table}
\centering
    \begin{tabular}{|l |c c c |c c c |}
    \hline
    & \multicolumn{3}{|c|}{\small{OOD}} & \multicolumn{3}{|c|}{\small{ID}}  \\
    OOD Clients & \small{Base }& \small{MoE} & \small{{\our}} & \small{Base}& \small{MoE} & \small{{\our}} \\
    \hline
    
    \small{Electronics+Games} & 86.9 & 87.4 & 87.7 & 88.6 & 88.7 &  89.0 \\
    \small{Industrial+Tools} & 87.6 & 88.3 &  87.7 & 88.4 & 88.8 &  88.9 \\
    \small{Books+Kindle Store} & 83.4 & 84.6 &  84.1 & 88.2 & 88.8 &  88.7 \\
    \small{CDs+Digital Music} & 82.4 & 83.0 &  83.2 & 89.0 & 88.9 &  88.9 \\
    \small{Arts+Automotive} & 90.2 & 90.6 &  90.4 & 88.4 & 88.6 &  88.6 \\
    \hline
    Average & 86.1 & 86.8  & 86.6 & 88.5 & 88.8 & \textbf{88.9} \\ \hline
    \end{tabular}
    \caption{Test Accuracy on Amazon Sentiment Data.}
    \label{table:sentiment_numbers}
\end{table}

    Table~\ref{table:sentiment_numbers} shows that average  OOD client accuracy increases from 86.1  to 86.8 with \our. 
    
    \noindent{\bf Auto-complete Task}
    We model this task as a forward language model 
    and measure perplexity. We used the 20 NewsGroup dataset and treat each of the twenty topics as a client. Thus $D$ is of size 20. 
    We use the state-of-art Mogrifier LSTM~\citep{MelisTP2020}. We perform leave-1-topic-out evaluation six times and OOD topics are shown in Table~\ref{tab:expts:lm}. For MoE, the client-specific parameter is only the bias and not the full softmax parameters which would blow up the number of trainable parameters. Also it did not perform well.
    \begin{table}[htb]
    \centering
    \begin{tabular}{|l|rrr|rrr|}
    \hline
    OOD & \multicolumn{3}{|c|}{\small{OOD}} & \multicolumn{3}{|c|}{\small{ID}}  \\
    Clients & Base & MoE & {\our} & Base & MoE & {\our}\\ \hline
    
    sci.space & 29.6 & 30.9 & 29.0 & 28.8 & 30.7 & 28.1\\
    comp.hw & 26.5 & 28.6 & 26.4 & 28.1 & 28.7 & 27.6\\
    sci.crypt & 29.7 & 29.8 & 29.6 & 27.8 & 28.1 & 27.7\\
    atheism & 28.3 & 28.1 & 28.1 & 27.9 & 28.2 & 28.0\\
    autos & 28.0 & 28.4 & 27.9 & 27.7 & 28.2 & 27.7\\
    mideast & 27.4 & 26.7 & 27.3 & 28.4 & 27.9 & 27.7\\ \hline
    Average & 28.2 & 28.7 & {\bf 27.9} & 28.0 & 28.8 & {\bf 27.7}\\ \hline
    \end{tabular}
    \caption{Perplexity comparison between the baseline and {\our} on 20-NewsGroup dataset.}
    \label{tab:expts:lm}
\end{table}
Table~\ref{tab:expts:lm} shows 
that \our\ performs consistently better than the baseline with average perplexity drop from 28.2 to 27.9. This drop is particularly significant because the Mogrifier LSTM is a strong baseline to start with. 
MoE is worse than baseline. 

\begin{figure}[ht] 
  \begin{subfigure}[b]{\linewidth}
    \centering
    \includegraphics[width=0.85\linewidth]{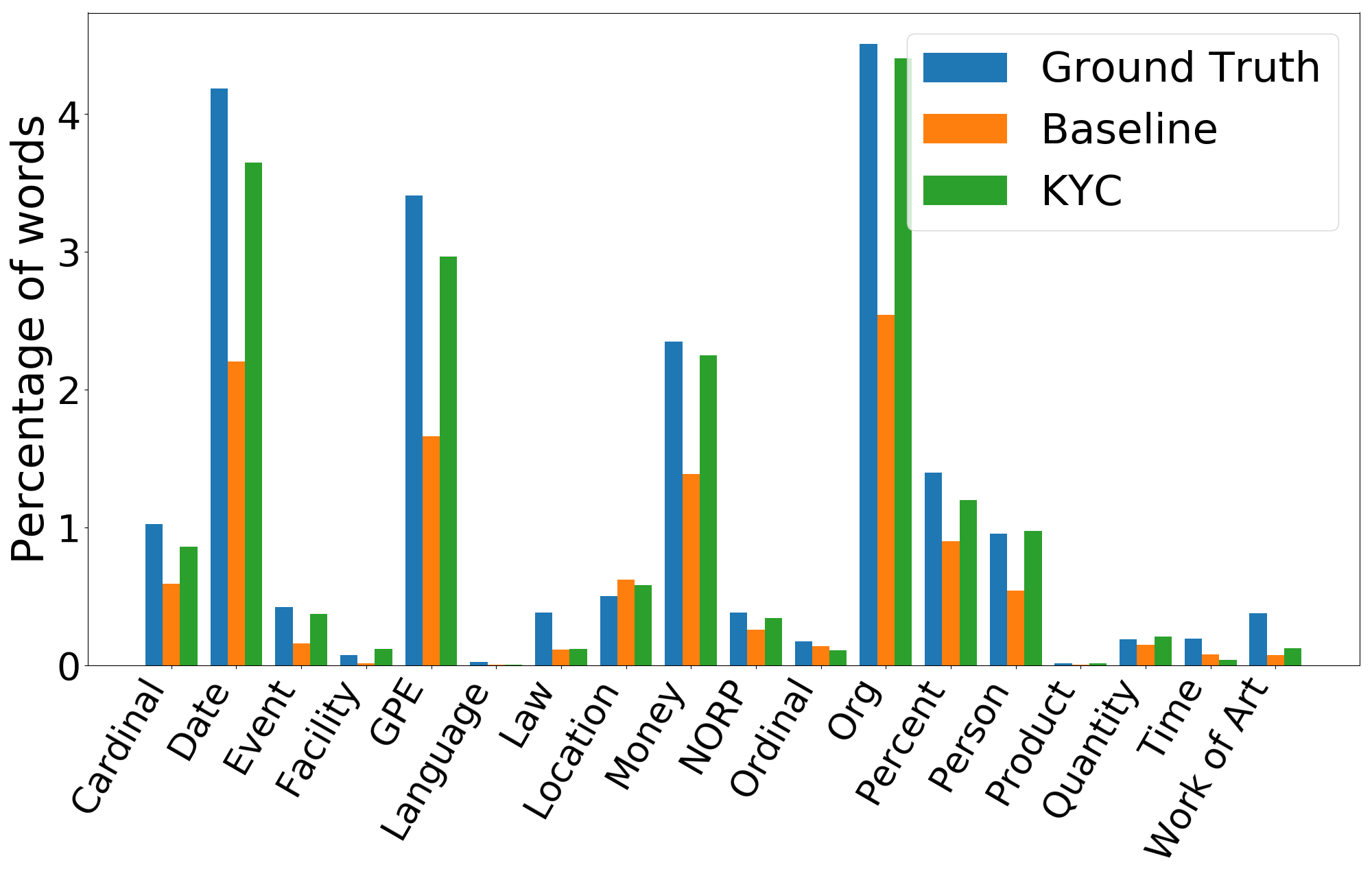} 
    \label{fig:bar_graph_xinhua} 
  \end{subfigure}

\caption{Proportion of true and predicted entity labels on OOD client NW/Xinhua.  Similar trends observed on other OOD domains~(Figure~\ref{fig:bar_graphs_more} of Appendix).}
  \label{fig:bar_graphs}
\end{figure}

\begin{figure}[ht] 
    \centering
    \includegraphics[width=.85\hsize]{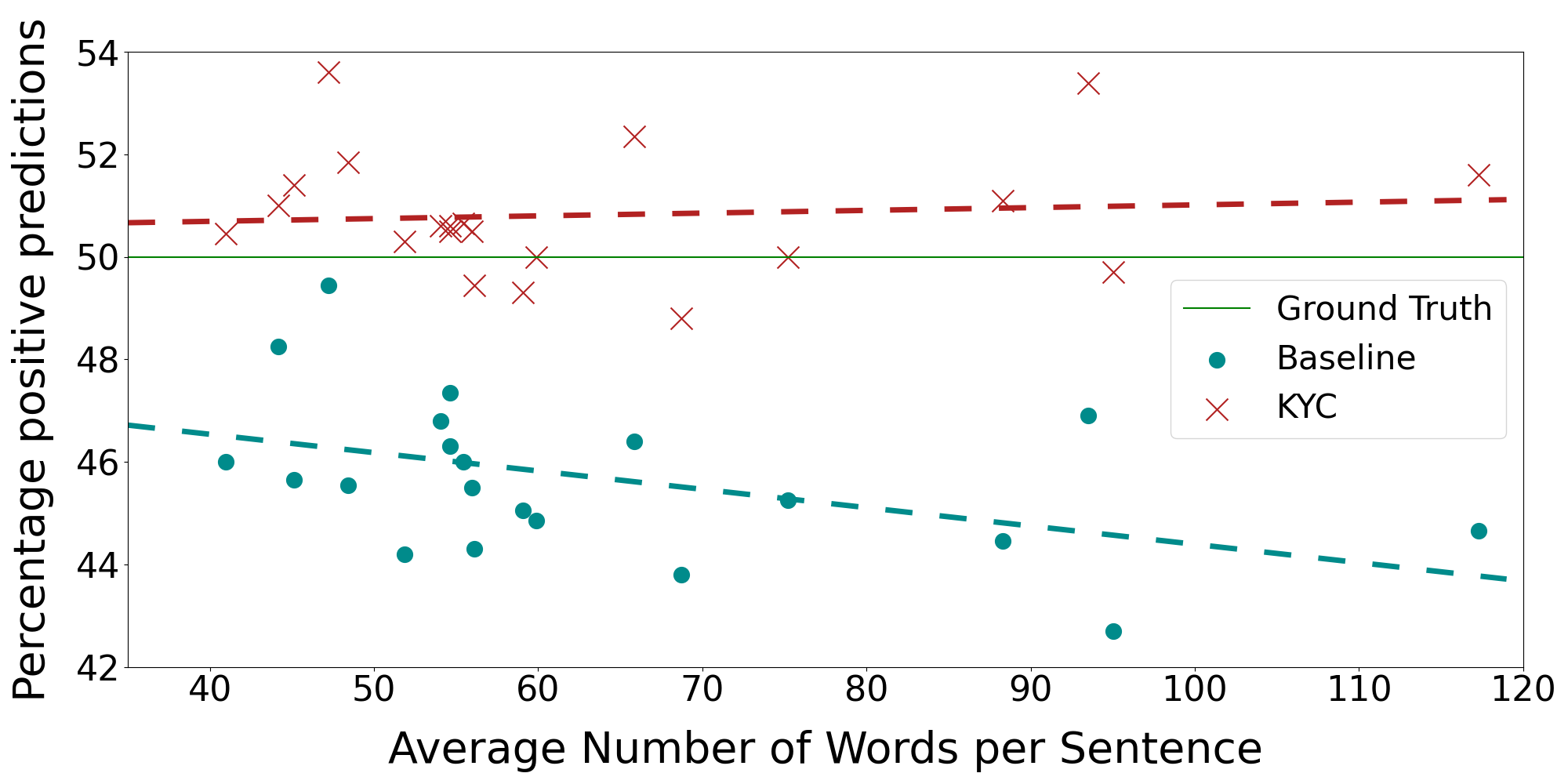} 
    \caption{Fraction Positive Predicted versus average review length by baseline and \our. Each dot/cross is a domain and the dotted lines indicate the best fit lines.} 
    \label{fig:sentiment_observation} 
\end{figure}

\paragraph{Statistical Significance}
We verify the statistical significance of the gains obtained for the Sentiment Analysis and Auto-complete tasks; the gains in the case of NER are much larger than statistical variation. Shown in Tables \ref{table:sentiment_stat_significance} and \ref{table:auto_comp_stat_significance} are the sample estimate and standard deviation for three runs along with the p value corresponding to the null hypothesis of significance testing. In both cases, we see that the gains of {\our} over the baseline are statistically significant.

\begin{table}[b]
\centering
\begin{tabular}{|l | c c c|}
\hline
    OOD Clients & \small{Base} & \small{{\our}} & \small{p-value} \\
    \hline
    \small{Electronics+Games} & 86.9(0.39) & 87.7(0.33) & 0.05\\
    \small{Industrial+Tools} & 87.6(0.19) & 87.7(0.09) & 0.14\\
    \small{Books+Kindle Store} & 83.4(0.03) & 84.1(0.14) & 0.01 \\
    \small{CDs+Digital Music} & 82.4(0.24) & 83.2(0.08) & 0.02 \\
    \small{Arts+Automotive} & 90.2(0.21) & 90.4(0.31) & 0.20\\
    \hline
    Average & 86.1(0.16) & 86.6(0.13) & 0.02\\ \hline
\end{tabular}
\vspace{0.3em}
\caption{Statistical significance of results on the OOD clients by {\our} for Sentiment Classification. For every entry contains the mean with the standard deviation in parenthesis}
\label{table:sentiment_stat_significance}
\end{table}

\begin{table}
\centering
\begin{tabular}{|l | c c c|}
\hline
    OOD Clients & \small{Base} & \small{{\our}} & \small{p-value} \\
    \hline
    sci.space & 26.5(0.4) & 26.4(0.2) & 0.39\\
    comp.hw & 29.6(0.4) & 29.0(0.3) & 0.07\\
    sci.crypt & 29.7(0.4) & 29.6(0.7) & 0.46 \\
    atheism & 28.3(0.2) & 28.1(0.2) & 0.14 \\
    autos & 28.0(0.5) & 27.9(0.4) & 0.34\\
    mideast & 27.4(0.4) & 27.3(0.4) & 0.37\\
    \hline
    Average & 28.2(0.2) & 27.9(0.0) & 0.04\\ \hline
\end{tabular}
\vspace{0.3em}
\caption{Statistical significance of results on the OOD clients by {\our} for the Auto Complete task. For every entry contains the mean with the standard deviation in parenthesis}
\label{table:auto_comp_stat_significance}
\end{table}

\paragraph{Diagnostics}
We provide insights on why \our's simple method of learning per-client label biases from client sketches is so effective.  %
One explanation is that the baseline had large discrepancy between the true and predicted class proportions for several OOD clients. \our\  corrects this discrepancy via {\em computed} per-client biases.  
Figure~\ref{fig:bar_graphs} shows true, baseline, and \our\ predicted class proportions for one OOD client on NER. Observe how labels like {\tt date}, {\tt GPE},  {\tt money} and {\tt org} are under-predicted by baseline and corrected by \our.
%
Since \our\ only corrects label biases, instances most impacted are those close to the shared decision boundary, and exhibiting properties correlated with labels but diverging across clients. We uncovered two such properties:

\noindent{\bf Ambiguous Tokens} In NER the label of several tokens changes across clients, E.g. tokens like {\tt million,} {\tt billion} in finance clients like NW/Xinhua are {\tt money} 92\% of the times whereas in general only 50\% of the times. 
Based on client sketches, it is easy to spot finance-related topics and increase the bias of {\tt money} label. This helps \our\ correct labels of borderline tokens.  

\noindent{\bf Instance Length}
For sentiment labeling, review length is another such property.  Figure~\ref{fig:sentiment_observation} is a scatter plot of the average review length of a client versus the fraction predicted as positive by the baseline. For most clients, review length is clustered around the mean of 61, but four clients have length $> 90$.  Length of review is correlated with label: on average, negative reviews contain 20 words more than positive ones.  This causes baseline to under-predict positives on the few clients with longer reviews.  The topics of the four outlying clients (video games, CDs, Toys\&Games) are related so that the client sketch is able to shift the decision boundary to correct for this bias. Using only normalized average sentence length as the client sketch bridges part of the improvement of {\our} over the baseline (details in Appendix C) implying that average instance length should be part of client sketch for sentiment classification tasks. 

\begin{table}
\centering
\begin{tabular}{| l | c | c c c | c c c|}
\hline
 & \small{Salience} & \small{TF} & \small{Binary} & \small{Sum-} & \multicolumn{3}{|c|}{Architecture} \\
 & \small{Concat} & \small{IDF} & \small{BOW} & \small{mary} & \small{Deep} & \small{Decomp} & \small{MoE-$\bm{g}$} \\
\hline
\small{OD} & 80.1 & 80.0 & 81.0 & 75.4 & 80.9 & 76.0 & 74.9  \\
\small{ID} & 86.0 & 85.9 & 77.8 & 81.8 & 85.9 & 85.0 & 79.8 \\
\hline
\end{tabular}
\vspace{0.3em}
\caption{Comparing variant client sketches ($S_c$) and network architectures ($\bigoplus$ and $Y_\theta$) of \our\ in Fig~\ref{fig:overview}.}
\label{table:ner_method_comparison}
\end{table}


\paragraph{Ablation Studies}
\label{sec:ablation}
We explored a number of alternative client sketches  and models for harnessing them.  We present a summary here; details are in the Appendix C and D.
Table~\ref{table:ner_method_comparison} shows average F1 on NER for three other sketches: TF-IDF, Binary bag of words, and a 768-dim pooled BERT embedding of ten summary sentences extracted from client corpus~\cite{gensim_summarizer}.  \our's default term saliency features provides  best accuracy with TF-IDF a close second, and embedding-based sketches the worst.
Next, we compare three other architectures for harnessing $\bm{g}$ in Table~\ref{table:ner_method_comparison}:
\textbf{Deep}, where module $\bigoplus$ after concatenating $\bm{g}$ and $\lastE$ adds an additional non-linear layer so that now the whole decision boundary, and not just bias, is client-specific. \our's OOD performance increases a bit over plain concat. 
\textbf{Decompose}, which mixes two softmax matrices with a client-specific weight $\alpha$ learned from $\bm{g}$.
\textbf{MoE-$\bm{g}$}, which is like MoE but uses the client sketch for expert gating.
We observe that the last two options are worse than \our. 

\section{Conclusion}

We introduced the problem of lightweight client adaption in NLP service settings.  This is a promising area, ripe for further research on more complex tasks like translation.  We proposed client sketches and \our: an early prototype server network for on-the-fly adaptation.  Three NLP tasks showed considerable benefits from simple, per-label bias correction.  Alternative architectures and ablations provide additional insights.

\bibliographystyle{acl_natbib}
\bibliography{main,ML,voila}

\clearpage
\twocolumn[\centering \bfseries \large \xtitle \\
(Appendix) \par\bigskip]
\appendix

\section{Reproducibility/Implementation Details}
In this section we provide details about the dataset, architecture and training procedures used for each of the three tasks.
We provide the datasets used, code, hyperparameters for all the tasks in the code submitted along with the submission.

\subsection{NER}
We use the standard splits provided in the Ontonotes dataset \cite{ontonotes}. Our codebase builds on the official PyTorch implementation released by \cite{devlin2018bert}. We finetune a cased BERT base model with a maximum sequence length of 128 tokens for 3 epochs which takes 3 hours on a Titan X GPU. 

\subsection{Sentiment Classification}
As described previously, we use the Amazon dataset \cite{amazon_dataset}. For each review, we use the standard protocol to convert the rating to a binary class label by marking reviews with 4 or 5 stars as positive, reviews with 1 or 2 stars as negative and leaving out reviews with 3 stars. We randomly sample data points from each domain to select 1000, 200 and 500 positive and negative reviews each for the train, validation and test splits, respectively. We leave out the domains that have insufficient examples, leaving us with 22 domains.
\indent We use the finetuning protocol provided by the authors of \cite{sentiment_repo} and use the uncased BERT base model with a maximum sequence length of 256 for this task. We train for 5 epochs (which takes 4 hours on a Titan X GPU) and use the validation set accuracy after every epoch to select the best model. 

\subsection{Auto Complete Task}
\label{sec:appendix:lm}
We use 20NewsGraoup dataset while regarding each content class label as a client. We remove header, footer from the content of the documents and truncate the size of each client to around 1MB. We use word based tokenizer with a vocabulary restricted to top 10,000 tokens and demarcate sentence after 50 tokens. The reported numbers in Table~\ref{tab:expts:lm} are when using TF-IDF vector for domain sketch. We diIn this section, we reportd not evaluate other kinds of domain sketch on this task. We train all the methods for 40 epochs with per epoch train time of 4 minutes on a Titan X GPU. 

We adopt the tuned hyperparameters corresponding to PTB dataset to configure the baseline~\citet{MelisTP2020}. Since the salience information from the client sketch can be trivially exploited in  perplexity reduction and thereby impede learning desired hypothesis beyond trivially copying the salience information, we project the sketch vector to a very small dimension of 32 before fanning it out to the size of vocabulary. We did not use any non-linearity in $G_\phi$ and also employ dropout on the sketches.

\section{Details of MoE method~\cite{GuoSB18}}
MoE employs a shared encoder and a client specific classifier.  We implemented their proposal to work with our latest encoder networks. Our implementation of their method is to the best of our efforts faithful to their scheme. The only digression we made is in the design of discriminator: we use a learnable discriminator module that the encoder fools while they adopt MMD based metric to quantify and minimize divergence between clients. This should, in our opinion, only work towards their advantage since MMD is not sample efficient especially given the small size of our clients.

\begin{figure*}[tb]
  \begin{subfigure}{0.5\linewidth}
   \centering
    \includegraphics[width=0.9\linewidth]{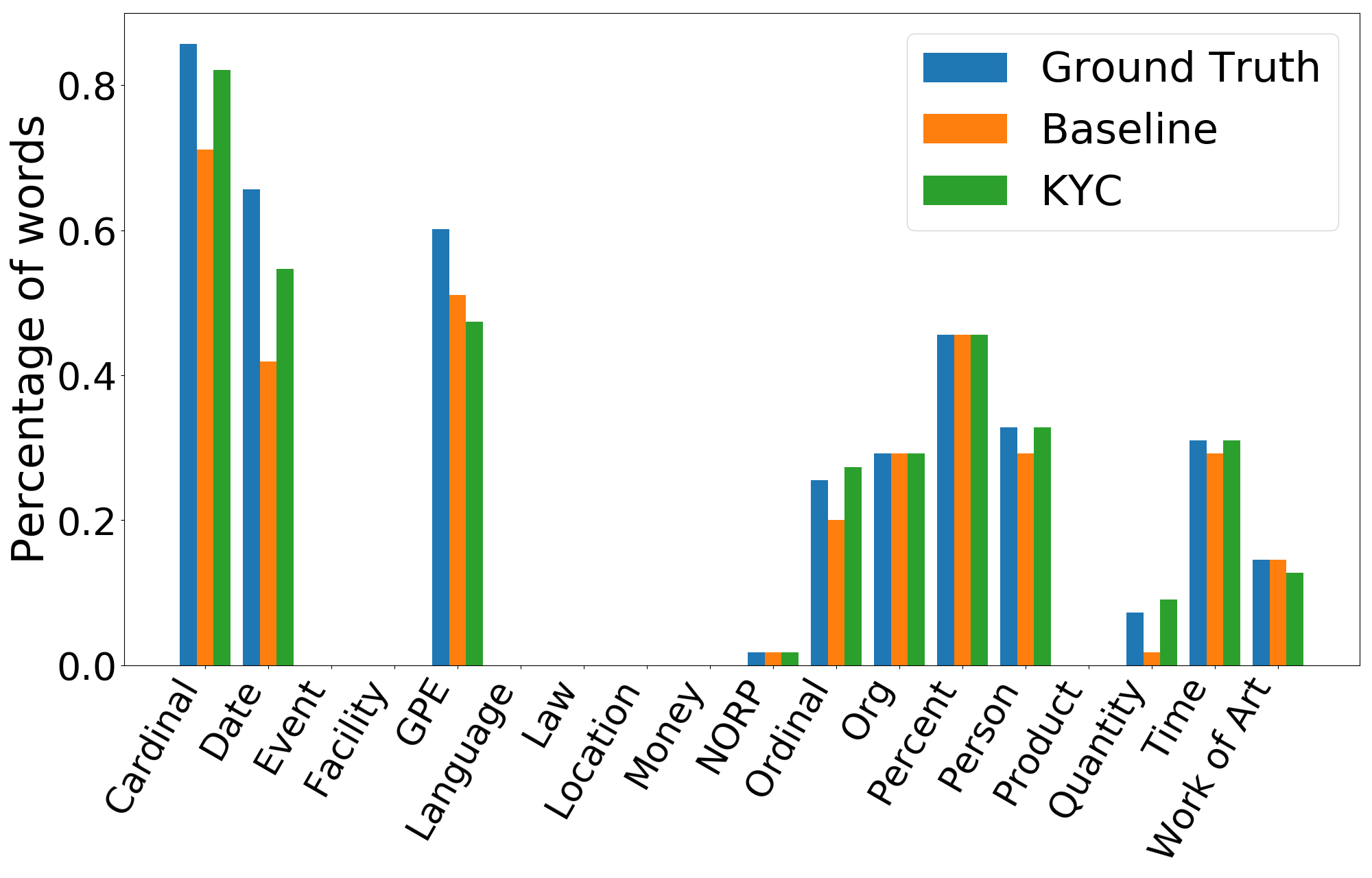} 
    \end{subfigure}
  \begin{subfigure}{0.5\linewidth}
  \centering
    \includegraphics[width=0.9\linewidth]{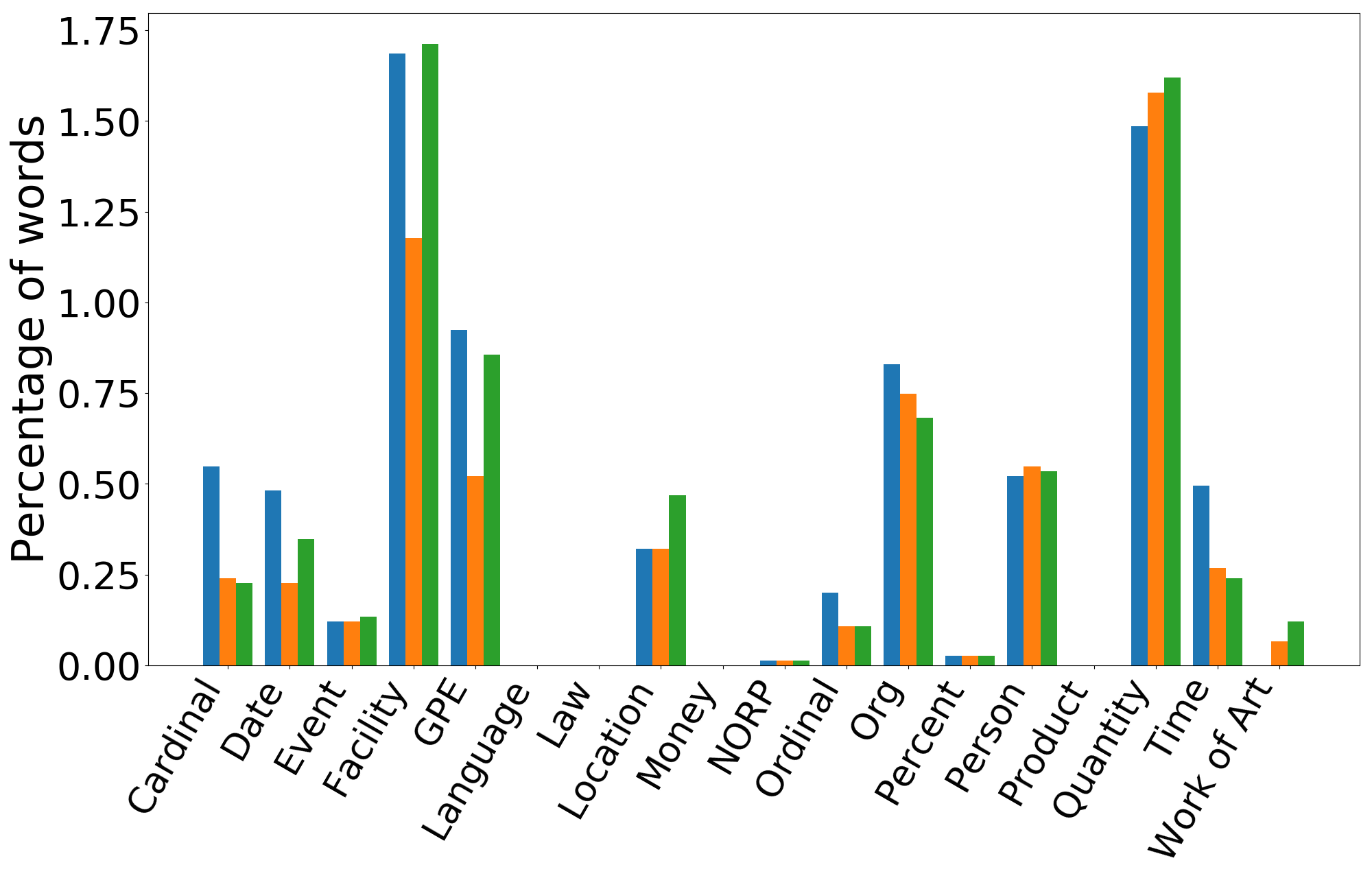} 
\end{subfigure}
\caption{Proportion of true and predicted entity labels for different OOD clients (left) BC/Phoenix (right) BC/CCTV.}
  \label{fig:bar_graphs_more}
\end{figure*}

\begin{table}[!htbp]
\centering
\begin{tabular}{|l | c c | c c|}
\hline
& \multicolumn{2}{|c|}{\small{OOD}} & \multicolumn{2}{|c|}{\small{ID}}  \\
OOD Clients & \small{Base}  & \small{{\our}} & \small{Base} & \small{{\our}} \\
\hline
\small{BC/CCTV + BC/Phoenix} & 63.8 & 70.1 & 86.00 & 86.7 \\
\small{BN/PRI + BN/VOA} & 88.7 & 91.6 & 84.5 & 86.2\\
\small{NW/WSJ + NW/Xinhua} & 73.9 & 79.2 & 80.8 & 82.2\\
\small{BC/CNN + TC/CH} & 78.3 & 80.4 & 85.6 & 87.1\\
\small{WB/Eng + WB/a2e} & 76.2 & 78.9 & 86.4 & 87.5 \\
\hline
Average & 76.2 & 80.0 & 84.7 & 85.9\\ \hline
\end{tabular}
\vspace{0.3em}
\caption{Performance on the NER task on the Ontonotes dataset when using TF-IDF as the client sketch.}
\label{table:ner_tfidf}
\end{table}

\begin{table}[!htbp]
\centering
\begin{tabular}{|l | c c | c c|}
\hline
& \multicolumn{2}{|c|}{\small{OOD}} & \multicolumn{2}{|c|}{\small{ID}}  \\
OOD Clients & \small{Base}  & \small{{\our}} & \small{Base} & \small{{\our}} \\
\hline
\small{BC/CCTV + BC/Phoenix} & 63.8 & 75.3 & 86.0 & 79.3 \\
\small{BN/PRI + BN/VOA} & 88.7 & 90.5 & 84.5 & 78.7\\
\small{NW/WSJ + NW/Xinhua}& 73.9 & 82.7 & 80.8 & 71.4\\
\small{BC/CNN + TC/CH} & 78.3 & 80.3 & 85.6 & 79.9\\
\small{WB/Eng + WB/a2e} & 76.2 & 76.4 & 86.4 & 79.6 \\
\hline
Average & 76.2 & 81.0 & 84.7 & 77.8 \\\hline
\end{tabular}
\vspace{0.3em}
\caption{Performance on the NER task on the Ontonotes dataset when using Binary Bag of Words as the client sketch.}
\label{table:ner_bloom}
\end{table}

\begin{table}
\centering
\begin{tabular}{|l | c c | c c|}
\hline
& \multicolumn{2}{|c|}{\small{OOD}} & \multicolumn{2}{|c|}{\small{ID}}  \\
OOD Clients & \small{Base}  & \small{{\our}} & \small{Base} & \small{{\our}} \\
\hline
\small{BC/CCTV + BC/Phoenix} & 63.8 & 61.5 & 86.0 & 83.0\\
\small{BN/PRI + BN/VOA} & 88.7 & 82.3 & 84.5 & 85.2 \\
\small{NW/WSJ + NW/Xinhua} & 73.9 & 82.3 & 80.8 & 75.0\\
\small{BC/CNN + TC/CH} & 78.3 & 72.5 & 85.6 & 83.2\\
\small{WB/Eng + WB/a2e} & 76.2 & 78.3 & 86.4 & 82.5\\
\hline
Average & 76.2 & 75.4 & 84.7 & 81.8\\\hline
\end{tabular}
\vspace{0.3em}
\caption{Performance on the NER task on the Ontonotes dataset when using sentence embddings averaged over an extracted summary.}
\label{table:ner_summary}
\end{table}

\begin{table}[!htbp]
\centering
    \begin{tabular}{|l |c c c |c c c |}
    \hline
    & \multicolumn{3}{|c|}{\small{OOD}} & \multicolumn{3}{|c|}{\small{ID}}  \\
    OOD Clients & \small{Base } & \small{Sali-} & \small{Avg} & \small{Base} & \small{Sali-} & \small{Avg} \\
     &  & \small{ence} & \small{Len} & & \small{ence} & \small{Len} \\
    \hline
    \small{Electronics+Games} & 86.4 & 88.1 & 86.9 & 88.5 & 89.0 & 88.6 \\
    \small{Industrial+Tools} & 87.4 & 87.6 & 88.3 & 88.2 & 88.9 & 88.8 \\
    \small{Books+Kindle Store} & 83.5 & 84.6 & 84.5 & 88.0 & 88.9 & 89.0 \\
    \small{CDs+Digital Music} & 82.5 & 83.0 & 83.1 & 89.0 & 89.0 & 89.0 \\
    \small{Arts+Automotive} & 89.9 & 90.6 & 90.2 & 88.2 & 88.6 & 88.5 \\
    \hline
    Average & 86.0 & 86.8  & 86.6 & 88.4 & 88.8 & 88.8 \\ \hline
    \end{tabular}
    \caption{Accuracy on the Sentiment Analysis task when using average review length as the client sketch. Columns ``Saliency" and ``Avg Len" refer to using {\our} with the default saliency features and normalized review lengths as client sketches, respectively.}
    \label{table:sentiment_length_numbers}
\end{table}

\section{Results with Different Client Sketches}
\label{appendix:domain_features}
In this section we provide results on every OOD split for the different client sketches described in Section \ref{sec:ablation} along with more details.
\begin{itemize}
    \item \textbf{TF-IDF}: This is a standard vectorizer used in Information Retrieval community for document similarity. We regard all the data of the client as a document when computing this vector.  
    The corresponding numbers using this sketch are shown in Table~\ref{table:ner_tfidf} and are only slightly worse than the salience features.
    \item \textbf{Binary Bag of Words (BBoW)}: A binary vector of the same size as vocabulary is assigned to each client while setting the bit corresponding to a word on if the word has occurred in the client's data.  
    We notice an improvement on the OOD set but a significant drop in ID numbers as seen in Table~\ref{table:ner_bloom},~\ref{table:ner_method_comparison}. We attribute this to the strictly low representative power of BBoW sketches compared to the other sketches. The available train data for NER is laced with rogue clients which are not labeled and are instead assigned the default tag: ``O''. Proportion of {\our}'s improvement on this task comes from the ability to distinguish bad clients and keeping their parameters from not affecting other clients. This, however, is not possible when the representative capacity of the sketch is compromised. Thereby we do worse on ID using this sketch but not on OOD meaning the model does worse on the bad clients (which are only part of ID, and not OOD). 
    \item \textbf{Contextualized Embedding of Summary}: We also experiment with using deep-learning based techniques to extract the topic and style of a client by using the ``pooled" BERT embeddings averaged over sentences from the client. Since the large number of sentences from every client would lead to most useful signals being killed upon averaging, we first use a Summary Extractor \cite{gensim_summarizer} to extract roughly 10 sentences per client and average the sentence embeddings over these sentences only. This method turns out to be ineffective in comparison to the other client sketches, indicating that sentence embeddings do not capture all the word-distribution information needed to extract useful correction.
    \item \textbf{Average Instance Length:} For the task of Sentiment Analysis, we also experiment with passing a single scalar indicating average review length as the client sketch in order to better understand and quantify the importance of average review length on the performance of {\our}. We linearly scale the average lengths so that all train clients have values in the range $[-1, 1]$. As can be seen in Table \ref{table:sentiment_length_numbers}, this leads to a significant improvement over the baseline.  In particular, the OOD splits  CDs + Digital Music and Books + Kindle Store have reviews that are longer than the average and consequently result in improvements when augmented with average length information. The gains from review length alone are not higher than our default term-saliency sketch indicating that term frequency captures other meaningful properties as well. 
\end{itemize}

\section{Results with Different Model Architectures}
\label{appendix:network_arch}
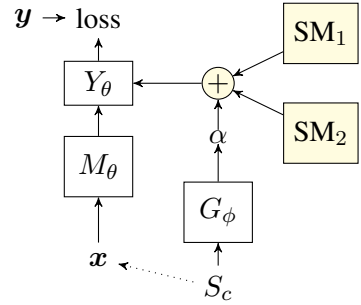
\begin{figure}
\begin{center}
\begin{tikzpicture}[>=stealth',align=center]
\node (loss) {loss};
\node [left=3mm of loss] (y) {$\bm{y}$};

\node [rectangle, draw,
minimum width=9mm, below=3mm of loss] (Ytheta) {$Y_\theta$};
\node [circle,draw, 
inner sep=.1mm, right=9mm of Ytheta, fill=yellow!15, ] (combination) {$+$};
\node [rectangle, draw, anchor=center,
minimum width=9mm, minimum height=8mm,
below=4mm of Ytheta] (Etheta) {$\lastE_\theta$};
\node [anchor=center,below=6mm of Etheta] (x) {$\bm{x}$};
\node [anchor=center,
below = 4mm of combination,
outer sep=.1mm, inner sep=0mm] (alpha) {$\alpha$};
\node [rectangle, draw, anchor=center,
minimum width=9mm, minimum height=8mm,
below=5mm of alpha] (Gphi) {$G_\phi$};
\node [rectangle, draw, anchor=center,
minimum width=9mm, minimum height=8mm,
above right=1mm and 7mm  of combination, fill=yellow!15, ] (sm1) {SM$_1$};
\node [rectangle, draw, anchor=center,
minimum width=9mm, minimum height=8mm,
below right=1mm and 7mm  of combination, fill=yellow!15, ] (sm2) {SM$_2$};
\node [anchor=center,below=3mm of Gphi] (sc) {$S_c$};
\draw [->] (y) to (loss);
\draw [->] (combination) to (Ytheta);
\draw [->] (sm2) to (combination);
\draw [->] (sm1) to (combination);
\draw [->] (alpha) to (combination);
\draw [->] (Ytheta) to (loss);
\draw [->] (Etheta) to (Ytheta);
\draw [->] (Gphi) to (alpha);
\draw [->] (x) to (Etheta);
\draw [->] (sc) to (Gphi);
\draw [->, dotted] (sc) to (x);
\end{tikzpicture}
\end{center}
\caption{Decompose overview: $\bigoplus$ indicates a weighted linear combination. SM$_i$, $i \in \{1,2\}$ represent the softmax matrices which are combined using weights $\alpha$.}  \label{fig:overview_decompose}
\end{figure}

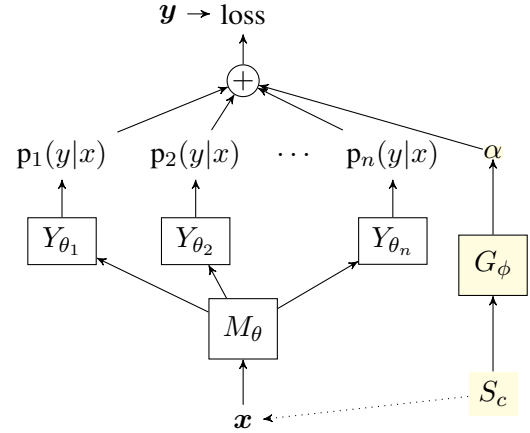
\begin{figure}[h]
\begin{center}
\begin{tikzpicture}[>=stealth',align=center]

\node (loss) {loss};
\node [left=3mm of loss] (y) {$\bm{y}$};
\node [circle,draw, 
inner sep=.1mm, below = 4mm of loss] (combination) {$+$};

\node [below left=5mm and 15mm of combination] (pred1) {p$_1(y|x)$};
\node [right=3mm of pred1] (pred2) {p$_2(y|x)$};
\node [right = 2mm of pred2] (dots) {$\ldots$};
\node [right=2mm of dots] (predn) {p$_n(y|x)$};
\node [rectangle, draw,
minimum width=9mm, below=5mm of pred1] (Ytheta1){$Y_{\theta_1}$};
\node [rectangle, draw,
minimum width=9mm, below=5mm of pred2] (Ytheta2){$Y_{\theta_2}$};
\node [rectangle, draw,
minimum width=9mm, below=5mm of predn] (Ythetan){$Y_{\theta_n}$};
\node [rectangle, draw, anchor=center,
minimum width=9mm, minimum height=8mm,
below=35mm of loss] (Etheta) {$\lastE_\theta$};
\node [anchor=center,
below right=7mm and 30mm of combination,
outer sep=.1mm, inner sep=0mm, fill=yellow!15, ] (alpha) {$\alpha$};
\node [rectangle, draw, anchor=center,
minimum width=9mm, minimum height=8mm,
below=10mm of alpha, fill=yellow!15, ] (Gphi) {$G_\phi$};
\node [anchor=center,below=10mm of Gphi, fill=yellow!15, fill=yellow!15, ] (sc) {$S_c$};
\node [anchor=center,below=6mm of Etheta] (x) {$\bm{x}$};
\draw [->] (x) to (Etheta);
\draw [->] (Etheta) to (Ytheta1);
\draw [->] (Etheta) to (Ytheta2);
\draw [->] (Etheta) to (Ythetan);
\draw [->] (Ytheta1) to (pred1);
\draw [->] (Ytheta2) to (pred2);
\draw [->] (Ythetan) to (predn);
\draw [->] (pred1) to (combination);
\draw [->] (pred2) to (combination);
\draw [->] (predn) to (combination);
\draw [->] (alpha) to (combination);
\draw [->] (Gphi) to (alpha);
\draw [->] (sc) to (Gphi);
\draw [->] (combination) to (loss);
\draw [->] (y) to (loss);
\draw [->, dotted] (sc) to (x);
\end{tikzpicture}
\end{center}
\caption{ MoE-$\bm{g}$ overview: $\bigoplus$ indicates a weighted linear combination. p$_i(y|x)$ represents the $i^{th}$ expert's predictions and $\alpha$ represents weights for expert gating.}  \label{fig:overview_moe}
\end{figure}

In this section we provide results for the different network architecture choices described in Section~\ref{sec:ablation}
\begin{itemize}
    \item \textbf{Deep}: 
    The architecture used is identical to that shown in Figure~\ref{fig:overview} barring $\bigoplus$, which now consists of an additional 128-dimensional non-linear layer before the final softmax transform $Y_\theta$.
    \item \textbf{Decompose}: The final softmax layers is decomposed in to two. A scalar $\alpha$ is predicted from the client sketch using $G_\phi$ similar to {\our}. The final softmax layer then is obtained through convex combination of the two softmax layers using $\alpha$. Figure~\ref{fig:overview_decompose} shows the overview of the architecture.
    \item \textbf{MoE-$\bm{g}$}: We use the client sketch as the drop-in replacement for encoded instance representation employed in \citet{GuoSB18}. 
    The architecture is sketched in  Figure~\ref{fig:overview_moe}. 
    As shown in Table~\ref{table:ner_moe}, this method works better than the standard MoE model, but worse than {\our}.
\end{itemize}

\begin{table}[!htbp]
\centering
\begin{tabular}{|l | c c | c c|}
\hline
& \multicolumn{2}{|c|}{\small{OOD}} & \multicolumn{2}{|c|}{\small{ID}}  \\
OOD Clients & \small{Base}  & \small{{\our}} & \small{Base} & \small{{\our}} \\
\hline
\small{BC/CCTV + BC/Phoenix} & 64.8 & 74.5 & 85.6 & 86.8 \\
\small{BN/PRI + BN/VOA} & 89.5 & 90.0 & 84.1 & 85.6\\
\small{NW/WSJ + NW/Xinhua} & 74.4 & 80.6 & 80.2 & 92.8\\
\small{BC/CNN + TC/CH} & 78.0 & 79.6 & 86.1 & 87.5 \\
\small{WB/Eng + WB/a2e}&   75.6 & 79.9 & 85.8 & 87.1\\
\hline
Average & 76.5 & 80.9 & 84.4 & 86.0 \\\hline
\end{tabular}
\vspace{0.3em}
\caption{Performance on the NER task on the Ontonotes dataset using KYC-Deep.}
\label{table:ner_deep}
\end{table}

\begin{table}[!htbp]
\centering
\begin{tabular}{|l | c c | c c|}
\hline
& \multicolumn{2}{|c|}{\small{OOD}} & \multicolumn{2}{|c|}{\small{ID}}  \\
OOD Clients & \small{Base}  & \small{{\our}} & \small{Base} & \small{{\our}} \\
\hline
\small{BC/CCTV + BC/Phoenix} & 64.1 & 56.0 & 85.6 & 86.3  \\
\small{BN/PRI + BN/VOA} & 89.6 & 89.9 & 84.6 & 85.5\\
\small{NW/WSJ + NW/Xinhua} & 72.3 & 68.2 & 81.2 & 80.0\\
\small{BC/CNN + TC/CH} & 78.5 & 77.5 & 85.9 & 86.6\\
\small{WB/Eng + WB/a2e} & 75.5 & 71.0 & 86.1 & 86.7 \\
\hline
Average & 76.0 & 72.5 & 84.7 & 85.2\\ \hline
\end{tabular}
\vspace{0.3em}
\caption{Performance on the NER task on the Ontonotes dataset using Decompose.}
\label{table:ner_softmax}
\end{table}

\begin{table}[!htbp]
\centering
\begin{tabular}{|l | c c | c c|}
\hline
& \multicolumn{2}{|c|}{\small{OOD}} & \multicolumn{2}{|c|}{\small{ID}}  \\
OOD Clients & \small{Base}  & \small{{\our}} & \small{Base} & \small{{\our}} \\
\hline
\small{BC/CCTV + BC/Phoenix} & 64.8 & 74.7 & 85.6 & 84.0 \\
\small{BN/PRI + BN/VOA} & 89.5 & 88.3 & 84.1 & 83.6\\
\small{NW/WSJ + NW/Xinhua} & 74.4 & 61.6& 80.2 & 64.8\\
\small{BC/CNN + TC/CH} & 78.0 & 73.7 & 86.1 & 82.1 \\
\small{WB/Eng + WB/a2e}&   75.6 & 76.3 & 85.8 & 84.4\\
\hline
Average & 76.5 & 74.9 & 84.4 & 79.8 \\\hline
\end{tabular}
\vspace{0.3em}
\caption{Performance on the NER task on the Ontonotes dataset using MoE-$\bm{g}$.}
\label{table:ner_moe}
\end{table}


\end{document}